\documentclass[11pt,a4paper]{article}
\usepackage[hyperref]{acl2020}
\usepackage{times}
\usepackage{latexsym}

\usepackage[normalem]{ulem}
\usepackage{booktabs}
\usepackage{colortbl}
\usepackage{tabularx}
\usepackage{graphicx}
\usepackage{tipa}
\usepackage[T1]{fontenc}
\usepackage[utf8]{inputenc}

\usepackage{microtype}

\aclfinalcopy 

\title{German Phoneme Recognition\\
with Text-to-Phoneme Data Augmentation}

\author{Dojun Park$^*$ \\
  Institute for Natural Language Processing \\
  University of Stuttgart \\
  \texttt{dojun.park@ims.uni-stuttgart.de} \\\And
  Seohyun Park$^*$ \\
  Department of German Language \\
   and Literature, Hankuk University\\
   of Foreign Studies \\
  \texttt{seohyun@hufs.ac.kr} \\}

\date{}

\begin{document}
\maketitle
\def\thefootnote{*}\footnotetext{These authors contributed equally to this work}
\begin{abstract}
In this study, we experimented to examine the effect of adding the most frequent n phoneme bigrams to the basic vocabulary on the German phoneme recognition model using the text-to-phoneme data augmentation strategy. As a result, compared to the baseline model, the vowel30 model and the const20 model showed an increased BLEU score of more than 1 point, and the total30 model showed a significant decrease in the BLEU score of more than 20 points, showing that the phoneme bigrams could have a positive or negative effect on the model performance. In addition, we identified the types of errors that the models repeatedly showed through error analysis.
\end{abstract}

\section{Introduction}

    With the advancement of AI, our society is changing rapidly. Machine translation, such as Google's translation service, has broken down language barriers and made it possible to exchange information quickly and widely. In addition, voice recognition systems such as Siri of Apple allow users to perform operations such as sending messages, searching for information, and setting up mobile phones using only their voice, replacing human effort in daily life. Deep learning, which is the basis of these AI technologies, shows a marked difference from previous machine learning approaches in that the machine extracts the characteristics of input information and information necessary for problem-solving from data by itself.
    
    From the perspective of deep learning, the development of phoneme recognition models faces considerable difficulties. Phoneme recognition refers to searching for the most probable phoneme sequence given a speech signal. In the case of speech recognition, which is being actively researched in the industry, it is not difficult to obtain a speech-text corpus for developing a speech recognition system. However, in the case of phoneme recognition, which has little interest in the industry and is relatively focused on academic research, it is difficult to obtain data constructed for phoneme recognition model training. 
    
    In this study, we propose a data augmentation strategy that extends the existing speech-text corpus to the speech-phoneme corpus to overcome these difficulties. In addition, we will take a closer look at whether the basic unit of vocabulary has a significant effect on the performance of phoneme recognition models.

\begin{table*}
    \resizebox{\textwidth}{!}{%
    \centering
    \arrayrulecolor{black}
    \begin{tabular}{ccccccccccc} 
    \toprule
    \textbf{epoch}~ & \textbf{base}~          & \textbf{vowel10}~ & \textbf{vowel20}~ & \textbf{vowel30}~       & \textbf{const10}~ & \textbf{const20}~       & \textbf{const30}~ & \textbf{total10}~ & \textbf{total20}~ & \textbf{total30}~        \\ 
    \midrule
    10~             & 22.28~                  & 21.95~            & 23.08~            & 22.64~                  & 20.60~            & 20.19~                  & 22.05~            & 23.76~            & 20.28~            & 11.57~                   \\ 
    20~             & 45.77~                  & 43.00~            & 44.93~            & 44.66~                  & 47.94~            & 45.70~                  & 46.60~            & 45.23~            & 43.20~            & 24.09~                   \\ 
    30~             & 48.17~                  & 46.45~            & 46.16~            & 47.34~                  & 46.93~            & 48.64~                  & 48.86~            & 45.82~            & 46.36~            & 25.69~                   \\
    40~             & 49.11~                  & 46.13~            & 48.57~            & 47.78~                  & 46.94~            & 47.51~                  & 49.17~            & 48.26~            & 47.87~            & 25.48~                   \\
    50~             & 49.81~                  & 50.28~            & 47.23~            & 50.47~                  & 50.84~            & 50.49~                  & 49.56~            & 50.88~            & 49.93~            & 27.90~                   \\
    60~             & 50.10~                  & 51.50~            & 48.38~            & 48.31~                  & 49.30~            & 51.17~                  & 51.23~            & 51.10~            & 50.54~            & 26.75~                   \\
    70~             & 50.89~                  & 50.63~            & 49.54~            & 51.92~                  & 51.09~            & 52.50~                  & \textbf{51.94}~   & 51.73~            & 51.82~            & \textbf{\uline{28.54}}~  \\
    80~             & \textbf{\uline{51.75}}~ & \textbf{51.68}~   & \textbf{51.23}~   & \textbf{\uline{53.33}}~ & 50.91~            & \textbf{\uline{53.11}}~ & 51.64~            & \textbf{52.08}~   & \textbf{52.17}~   & 28.10~                   \\
    90~             & 50.03~                  & 50.90~            & 50.85~            & 51.85~                  & 50.89~            & 51.44~                  & 51.90~            & 50.95~            & 50.81~            & 27.91~                   \\
    100~            & 50.59~                  & 50.67~            & 50.14~            & 51.23~                  & \textbf{51.22}~   & 52.31~                  & 51.82~            & 51.27~            & 50.51~            & 27.57~                   \\
    \bottomrule
    \end{tabular}
    }
    \arrayrulecolor{black}
    \caption{\label{font-table}BLEU scores for each model measured every 10 epochs.}
    \end{table*}

\section{Methods}

\subsection{Text-to-Phoneme Data Augmentation}

    \begin{figure}[t]
      \centering
      \includegraphics[width=\linewidth]{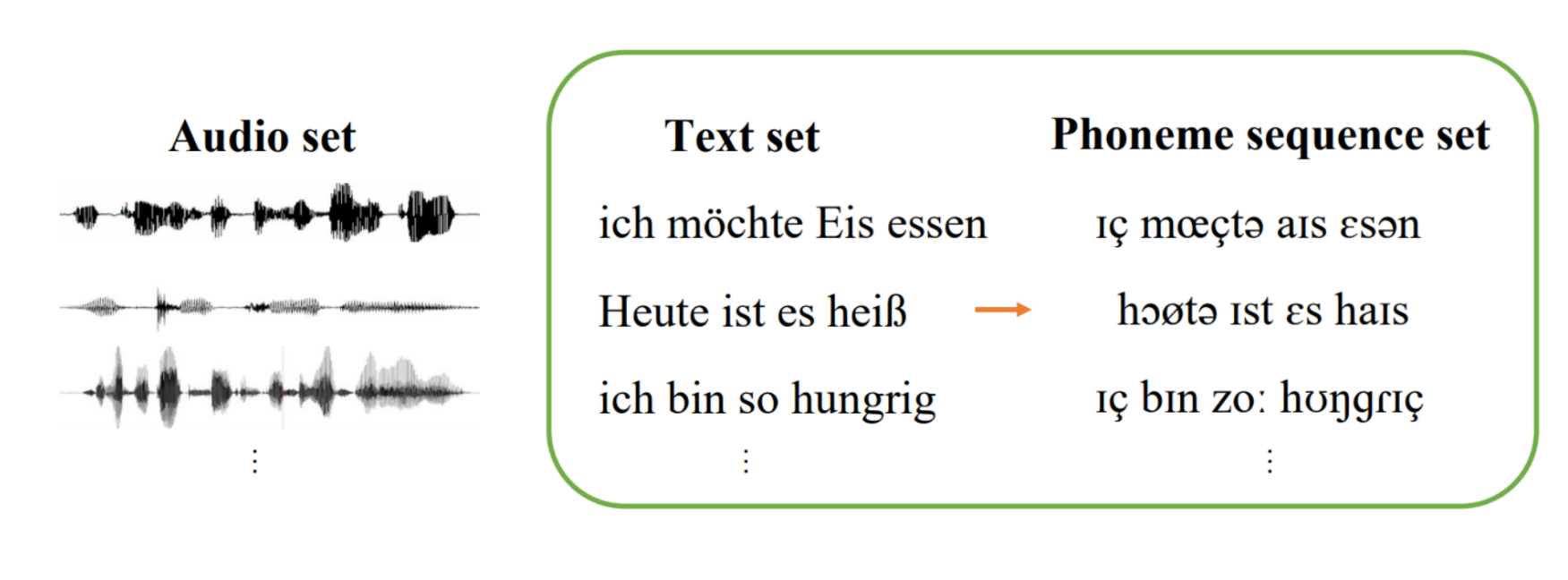}
      \caption{Data Augmentation with Text-to-Phoneme Transformation.}
      \label{fig:text-to-phoneme}
    \end{figure}

    figure 1 shows the text-to-phoneme transformation process to obtain the audio-phoneme parallel corpus, which is the data set of our study, from the audio-text parallel corpus. Producing a refined speech-phoneme corpus consumes a lot of time and money in that it requires manual annotation work by skilled experts. In addition, it is not easy to obtain speech data with phoneme annotations, and even if it is obtained, it is difficult to obtain a sufficient amount of data to be used as training data for machine learning. On the other hand, it is relatively easier to obtain a refined speech-text parallel corpus. Therefore, we intend to construct a speech-phoneme parallel corpus by performing data augmentation using the existing speech-text parallel corpus.
    
    We used epitran \cite{mortensen-etal-2018-epitran} for the text-to-phoneme data augmentation. Epitran is a Python library that supports text-to-phoneme conversion models for over 130 languages. We use this library as a connection to extend the speech-phoneme corpus from the speech-text corpus, and then use it as a dataset for our model training to develop phoneme recognition models.

\subsection{Introducing the most frequent bigrams into vocabulary}

    We changed the training conditions by adding the n most frequent phoneme bigrams to the vocabulary constituting the basic unit of the model output. First, we trained a baseline model with 49 single phonemes as a basic vocabulary. Next, 3 models with the top 10, 20, and 30 vowel bigrams were added as basic vocabulary, and the top 10, 20, and 30 consonant bigrams including the basic vocabulary were additionally trained. Finally, we trained the other three models by adding the top 10, 20, and 30 frequent phoneme pairs to the basic vocabulary from all phonemes without distinction between consonants and vowels. Accordingly, we trained 10 models with different vocabulary sizes, and we want to investigate the effect of the different settings in vocabulary on model performance through experiments.

\section{Experiment}

\subsection{Experimental setup}
    We used the German speech corpus of CSS10 \cite{park19c_interspeech} for this experiment. CSS10 is a single-speaker speech dataset consisting of 10 languages, and speech audio signals segmented in sentence units are mapped to text sentences in each language. We created a speech-phoneme corpus for our phoneme recognition modeling by converting German texts in this corpus into phoneme sequences using epitran. After excluding two sets exceeding 200 characters, we obtained 7425 speech-phoneme sequence pairs. We experimented by assigning 6425 pairs to the training set, and 500 pairs to the validation set and test set, respectively.
    
    We used the Transformer \cite{NIPS2017_3f5ee243} for our model training. For a fair comparison, the model training was performed by setting the same hyperparameters for each model. We set the number of hidden layers to 200, the number of heads to 2, the feed-forward layer to 400, the number of encoder layers to 4, and the number of decoder layers to 1. We trained the model by repeating 100 epochs and measured the training loss and validation loss for each epoch to track the model training progress. In addition, to verify the actual performance of the model, the model weights were stored at every 10th epoch point. 
    
    The performance of models was measured using BLEU \cite{papineni-etal-2002-bleu}. BLEU is a metric that has traditionally been widely used to measure the performance of models in machine translation. Unlike word error rate (WER), which measures accuracy at the character level based on the edit distance, BLEU considers the precision of character sequences from 1-gram to 4-gram. In this regard, we judged that the BLEU metric is more suitable than WER for measuring the performance of this experimental model, which must consider the precision of phoneme sequences. For BLEU measurement, we used the corpus\_bleu function of the nltk library. We calculated the geometric mean by applying the same weight of 0.25 to the modified precision of 1-gram to 4-gram. Then, if the predicted phoneme sequence is shorter than the correct phoneme sequence, a brevity penalty is applied to calculate the final BLEU score.

\subsection{Training}
    In figure 2, we calculate the loss of our 10 models: a baseline model with only a single phoneme as a basic vocabulary, models that add the top 10, 20, and 30 frequent bigrams to the basic vocabulary for each consonant, vowel, and all phonemes. In this way, we track the training progress. The left figure is the loss figure measured with the training set, and the right figure is the loss figure measured with the validation set. 
    
    First, looking at the training loss figure on the left, the loss falls sharply from 1 to 3 epochs, showing a fast learning process. After that, it shows a rather slow learning trend by drawing a gentle descending curve from 4 to 10 epochs, and then again, it is observed that the loss decreases rapidly until 20 epochs. Then, the descending curve gradually becomes gentle, but it can be seen that the training proceeds by continuously reducing the loss. 
    
    Next, looking at the validation loss on the right, it is observed that training proceeds similarly to the training loss figure until the 20 epochs. However, unlike the continuous decrease in training loss after 20 epochs, validation loss stagnates and increases slightly after 25 epochs. This suggests that the model has difficulty learning when the model is evaluated with the validation set not included in the training set because overfitting the training set occurs during model training.
    \begin{figure}[t]
      \centering
      \includegraphics[width=\linewidth]{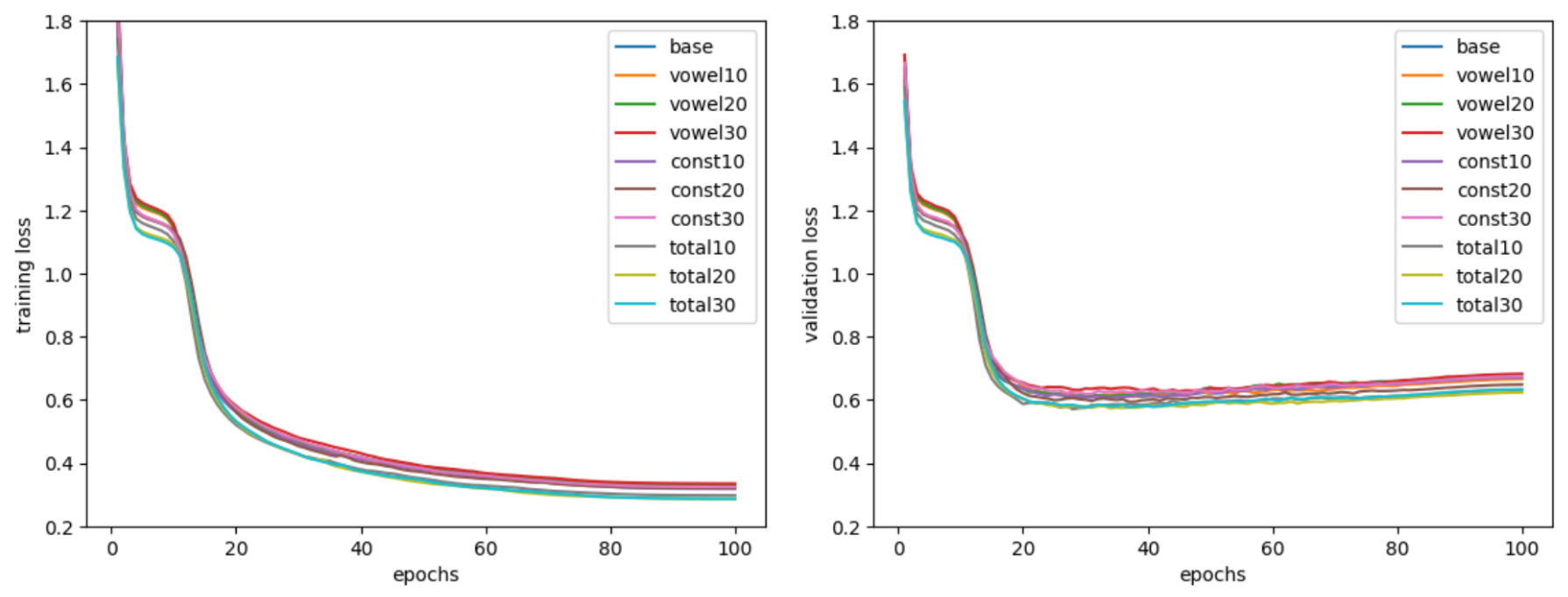}
      \caption{The training loss(left) and the validation loss(right).}
      \label{fig:loss}
    \end{figure}

\subsection{Results}

    \begin{table*}
    \resizebox{\textwidth}{!}{%
    \centering
    \arrayrulecolor{black}
    \begin{tabular}{cl} 
    \toprule
    \multicolumn{1}{l}{\textbf{Model}~} & \textbf{Text/phoneme sequence}~                                                                                                                                \\ 
    \midrule
    German~                                                                         & als sie von dem schönen Geist und dem Bartscherer überfallen wurden~                                                                                           \\
    Answer~                                                                         & \textipa{\textbf{als si: fo:n de:m S\o:n@n gaIst Und de:m baR\texttoptiebar{t\textesh}@R@R y:b@Rfal@n vURd@n}}~                                                                                 \\
    Baseline~                                                                       & \textipa{\textbf{als si: fo:n de:m b}A:\textbf{R y:b@Rfal@n v}\textopeno \textbf{Rd@n} als si: fo:n de:m bA:R fal@n v\textopeno R    d@n}~                                                       \\
    vowel10~                                                                        & \textipa{\textbf{als i: fo:n de:m b\o:s@n gaIst Und de:m} b\o:s@n gaIst Und de:m \textbf{b}A:\textbf{R S}e:\textbf{R@R y:b@Rfal@n v}\textopeno \textbf{Rd@n}}~                       \\
    vowel20~                                                                        & \textipa{\textbf{als si: fo:n de:m} bA:R fal@n v\textopeno  R   d@n \textbf{gaIst Und de:m b}A:\textbf{R fal@n v}\textopeno  \textbf{Rd@n}}~                                                      \\
    vowel30~                                                                        & \textipa{\textbf{als si: fo:n de:m} b\textbf{\o:}s\textbf{@n gaIst Und de:m baRtS@R@}n \textbf{v}\textopeno  \textbf{Rd@n}}~                                                         \\
    const10~                                                                        & \textipa{\textbf{als si: fo:n de:m} baRd@n als si: fo:n \textbf{de:m} ba\textbf{Rd@n}}~                                                                                  \\
    const20~                                                                        & \textipa{\textbf{als si: fo:n de:m} b\textbf{\o:}s\textbf{@n gaIst Und de:m} b\o:s@n gaIst Und de:m \textbf{b}A:\textbf{t S@R}I\texttoptiebar{t\textesh}t Und de:m b\o:s@n gaIst Und de:m bA:t}~  
                          \\
    const30~                                                                        & \textipa{\textbf{als si: fo:n de:m} b\textbf{\o:}s\textbf{@n gaIst Und de:m} vA:\textbf{R fal@n v}\textopeno  \textbf{Rd@n}}~                                                        \\
    total10~                                                                        & \textipa{\textbf{als si: fo:n de:m} b\textbf{\o:s@n gaIst Und de:m} b\o:s@n gaIst Und de:m \textbf{baRt S}E:\textbf{R y:bfal@n v}\textopeno  \textbf{Rd@n}}~                          \\
    total20~                                                                        & \textipa{\textbf{als si: fo:n de:m} b\textbf{\o:}s\textbf{@n gaIst Und de:m} b\o:s@n gaIst Und de:m b\o:s@n gaIst Und de:m b\o:s@n gaIst Und de:m}~                          \\
    total30~                                                                        & \textipa{e:ss au \textbf{fo:n} st\textbf{m b}A:t SstSstdnd \textbf{v}\textopeno  \textbf{Rd}nd}~                                                                                    \\
    \bottomrule
    \end{tabular}
    }
    \arrayrulecolor{black}
    \caption{predictions of each phoneme recognition model including German text and answer.}
    \end{table*}
    
    We can check the BLEU scores of each model that has been trained to 100 epochs in Table 1. We saved the trained weights every 10 epochs for 10 training models, and the predicted output by each model for the test speech signal was measured as a BLEU score. As training progresses, a gradual increase in BLEU scores is observed in all models. Unlike the validation loss figure, where the model did not train after 20 epochs, the BLEU scores measured based on the actual phoneme sequence prediction of the model progress gradually to 70-80 epochs in most models. The BLEU scores, which peaked thereafter, drop slightly after 90 epochs. The reason for this is that the actual model training reached the optimum point at 70 to 80 epochs, and then, the performance of the test set deteriorated due to the overfitting of the training data. However, the const10 model, in which the 10 most frequent consonant bigrams are added to the vocabulary, shows the BLEU score of 50.91 at 70 epochs and 50.91 and 50.89 at 80 and 90 epochs, which seemed to gradually decrease. However, it shows 51.22 at 100 epochs, which is 0.33 higher than 90 epochs, the highest BLEU score of the model.
    
    The base model, which trained only a single phoneme as a basic vocabulary, showed 51.75 at 80 epochs, showing the highest BLEU score in this model, and other models generally showed similar learning patterns. However, the vowel30 model, which added 30 of the most frequent vowel bigrams as a basic vocabulary, showed 53.33 at 80 epochs, which is the highest BLEU score among all models, showing a BLEU score 1.58 higher than the base model. In addition, the const20 model, which added 20 bigrams of the most frequent consonant bigrams to the basic vocabulary, also showed the highest BLEU score of 53.11 at the 80 epoch, which is 1.36 higher than the base model. Contrastively, the model in which 30 bigrams of the most frequent phoneme bigrams including both consonants and vowels were added showed a BLEU score of 11.57 at 10 epochs, which was only half the performance of other models with a BLEU score of 20 or higher at the same epochs. The BLEU score of this model gradually increased from 10 epochs and reached 28.09 at 80 epochs, the highest BLEU score among this model, which is 23.66 points lower than the base model.
    
    This result shows that the type and number of the vocabulary of the phoneme recognition model have a significant effect on model training. That vowel30 and const20 models showed BLEU scores that increased by more than 1 point compared to the baseline model indicates that the model performance can be improved by adding appropriate vowel and consonant bigrams to the vocabulary. However, the addition of inappropriate bigrams does not have a significant effect on model training considering that only models that added a certain number of vocabulary could induce a significant performance improvement of more than 1 point. Looking at the BLEU score of the Total30 model, it is observed that the vocabulary of the phoneme bigrams that do not distinguish between consonants and vowels actually caused confusion in model training, and thus showed low predictive performance. This suggests that setting phonetically insignificant bigrams as the default vocabulary can degrade the performance of the model.

\section{Error Analysis}

\subsection{Sentence analysis}
    
        Table 2 shows the German text and correct answer phoneme sequence along with the phoneme recognition prediction results of the model at the epoch point with the highest performance as a result of model training. To distinguish the correct predicted phoneme sequence, we marked the predicted phoneme that matches the answer among the prediction results in bold. 

        Most models show prediction results consistent with the correct answer for the phoneme sequence “als si: fo:n de:m” of the first four words. However, the model vowel10 dropped the “s” of “si:”, and the total30 model output a different phoneme sequence “e:ss au”. All models failed to predict the following phoneme sequence "\textipa{S\o:n@n}" correctly, but vowel10, vowel30, const20, const30, total10, and total20 models predicted a similar sequence "\textipa{b\o:s@n}". Next, for the phoneme sequence "\textipa{gAIst Und de:m}", all models except the base, const10, and total30 showed predictions consistent with the correct answer, and the base and const10 models accurately predicted only a partial “de:m”. For the phoneme sequence “\textipa{bA\;R\texttoptiebar{t\textesh}@\;R@\;R}”, all models failed to predict correctly, and only vowel10, vowel30, const20, and total10 models predicted some matching sequences, such as “\textipa{bA:\;R Se:\;R@\;R}”, “\textipa{bA\;RtS@\;R@n}”, “\textipa{bA:t S@\;RI\texttoptiebar{t\textesh}t}”. For the phoneme sequence “\textipa{y:b@Rfal@n}”, only the vowel10 model showed a correct prediction, but the total10 model showed a prediction close to the answer, showing “\textipa{y:bfal@n}“ with missing “\textipa{@R}”. Baseline, vowel20, and const30 only predict “\textipa{fal@n}”, which is part of the sequence. For the last phoneme sequence “\textipa{vURd@n}”, the baseline, vowel10, vowel20, vowel30, const30, and total10 models predicted “\textipa{v\textopeno   Rd@n}" most similar to the answer, the const10 model predicted “\textipa{baRd@n}” and total30 predicted “\textipa{v\textopeno  Rdnd}". 
        
        We were able to discover repeated error patterns through test sentence analysis. First, a phenomenon in which a specific place of the correct phoneme sequence is repeated is observed. Among the prediction results of the baseline, the phoneme sequence “\textipa{als si: fo:n de:m bA:R fal@n v\textopeno  Rd@n}” after the bold font indicating the appropriate prediction is a representative example. It is judged that this is caused by overfitting without reaching sufficient knowledge required for model prediction during the model training process. The next observed phenomenon is phoneme dropout which occurs at a specific location. This can be seen in the first two words “als i:” predicted by the vowel10 model. This model had to predict “als si:”, but only “i:” was returned as a prediction result while omitting the “s” of “si:”. This is considered to have added confusion to the successive phoneme “s” predicting the second phoneme “s” appropriately. Also, in the case where \textipa{vURd@n}, the phoneme sequence of the last word in the answer, was predicted as \textipa{v\textopeno  Rd@n}, it was confirmed that \textipa{U} was predicted as another similar vowel phoneme \textipa{\textopeno }. This shows the tendency of the model to return predictions with phoneme tokens similar to the correct answer, even if the predicted phoneme does not match the correct answer. 

\subsection{Definite Article Analysis}
    \begin{table}
    \small
    \arrayrulecolor{black}
    \resizebox{\columnwidth}{!}{%
    \begin{tabularx}{\columnwidth}{p{0.9cm}p{0.46cm}p{0.46cm}p{0.46cm}p{0.46cm}p{0.46cm}p{0.46cm}p{0.46cm}}
    \toprule
    \textbf{Model}~ & \textbf{der}~ & \textbf{des}~ & \textbf{dem}~ & \textbf{den}~ & \textbf{die}~ & \textbf{das}~ & \textbf{Avg.}~  \\ 
    \midrule
    base~            & 0.54~         & 0.53~         & 0.49~         & 0.63~         & 0.55~         & 0.53~         & 0.545~             \\
    vowel10~         & 0.47~         & 0.49~         & 0.54~         & 0.54~         & 0.56~         & 0.52~         & 0.520~             \\
    vowel20~         & 0.51~         & 0.46~         & 0.55~         & 0.65~         & 0.62~         & 0.60~         & 0.564~             \\
    vowel30~         & 0.62~         & 0.47~         & 0.56~         & 0.55~         & 0.70~         & 0.54~         & \textbf{0.573}~    \\
    const10~         & 0.53~         & 0.40~         & 0.58~         & 0.46~         & 0.63~         & 0.54~         & 0.523~             \\
    const20~         & 0.61~         & 0.61~         & 0.44~         & 0.58~         & 0.68~         & 0.57~         & \textbf{0.583}~    \\
    const30~         & 0.49~         & 0.39~         & 0.58~         & 0.57~         & 0.64~         & 0.56~         & 0.538~             \\
    total10~         & 0.58~         & 0.58~         & 0.46~         & 0.46~         & 0.54~         & 0.60~         & 0.536~             \\
    total20~         & 0.51~         & 0.49~         & 0.48~         & 0.54~         & 0.59~         & 0.53~         & 0.526~             \\
    total30~         & 0.50~         & 0.42~         & 0.43~         & 0.46~         & 0.59~         & 0.55~         & \textbf{0.492}~    \\
    \bottomrule
    \end{tabularx}
    }
    \arrayrulecolor{black}
    \caption{Accuracy of the definite articles.}
    \end{table}
    
    In Table 3, we measured the accuracy of how much the definite article of the answer sentences agrees with the prediction only for the definite article that appears frequently in the corpus. The mean of all models lies between 0.492 and 0.583. First, the const20 model that added the 20 most frequent consonant pairs to the basic vocabulary with the highest precision showed 0.583, and the vowel30 model that added the 30 most frequent vowels to the basic vocabulary showed 0.573. These two models showed the highest performance when measured by the BLEU score, and improved performance compared to the base model is observed not only in the BLEU score but also in the accuracy of the definite articles. The Total 30 model showed 0.492, which was less than 0.5, and showed the lowest precision value among the experimental models. This model also showed the lowest performance in performance comparison through the BLEU score, which is also observed again by measuring the accuracy of the definite articles. 

\section{Conclusions}

    We experimented to examine the effect of vocabulary on the performance of a German phoneme recognition model trained with a speech-phoneme corpus created using a German speech-text corpus. As a result of measuring the performance of the model with the BLEU metric, the vowel30 model showed 53.33 and the const20 model 53.11, which are 1.58 and 1.36 improved values compared to the baseline model, respectively. The total30 model showed a big performance decrease of 23.21 compared to the baseline model. These results suggest that adding appropriate bigrams to the basic vocabulary can bring meaningful improvement in model performance, while inappropriate phoneme pairs can have a significant negative impact on model performance. Through the phoneme sentence analysis, repeated error patterns were observed, such as an overfit error of repeatedly outputting a specific phoneme sequence, an error of outputting consecutive identical phonemes as one phoneme, and an error of predicting a specific vowel as another similar vowel. As a result of measuring the accuracy by limiting the analysis target to the definite article, the vowel30 model and the const20 model showed 0.028 and 0.038 improved accuracy compared to the baseline model, respectively, and the total30 model showed a 0.053 decrease in performance, which is consistent with the model performance analysis measured with BLEU. However, there is a limitation that the speech-phoneme parallel corpus of 6425 sentences used as training data for this experiment is not sufficient for deep learning model training. Therefore, it is considered that a follow-up study using a large-capacity speech corpus is necessary to prove the validity of this experimental result.

\bibliography{acl2020}
\bibliographystyle{acl_natbib}

\end{document}